# TONGUE CONTOUR EXTRACTION FROM ULTRASOUND IMAGES BASED ON DEEP NEURAL NETWORK

*Aurore Jaumard-Hakoun[1,2*], Kele Xu[1,2*], Pierre Roussel-Ragot[2*], Gérard Dreyfus[2], Maureen Stone[3], Bruce Denby[1,2*]*

[1]Université Pierre et Marie Curie, Paris, France
[2] SIGnal processing and MAchine learning Lab, ESPCI ParisTech, PSL Research University, Paris, France
[3]Vocal Tract Visualization Lab, University of Maryland Dental School, Baltimore, USA
*Present affiliation: Institut Langevin, ESPCI ParisTech, PSL Research University, Paris, France
`aurore.hakoun@espci.fr, denby@ieee.org`

**ABSTRACT**

Studying tongue motion during speech using ultrasound is a standard procedure, however automatic ultrasound image labelling remains a challenge, as standard tongue shape extraction methods typically require human intervention. This article presents a method based on deep neural networks to automatically extract tongue contours from speech ultrasound images. We use a deep autoencoder trained to learn the relationship between an image and its related contour, so that the model is able to automatically reconstruct contours from the ultrasound image alone. We use an automatic labelling algorithm instead of time-consuming hand-labelling during the training process. We afterwards estimate the performances of both automatic labelling and contour extraction as compared to hand-labelling. Observed results show quality scores comparable to the state of the art.

## 1. INTRODUCTION

Although ultrasound (US) provides a non-invasive and easy to implement tongue imaging method, the presence of multiplicative (Rayleigh) noise makes contour extraction with standard image processing techniques a challenge. Currently, most tongue contour extraction algorithms augment raw image data with a priori knowledge based on the physics of tongue movement. Simple examples require that contours found in a given frame be spatially "smooth" or forbid abrupt changes in contour shape between consecutive frames.

In [1], it has been shown that a deep neural network architecture is able to learn the contour extraction task when trained on hand-labelled contours. In this case, the smoothness criterion arises naturally because hand labelling is guided by a priori knowledge of the class of forms that a human tongue can assume. Hand labelling, however, is time consuming, and, furthermore, does not provide an obvious means of including the second constraint, i.e., that contours extracted from frames nearby in time must be "similar".

In this article, we repeat the procedure of [1], but replace hand-labelled training data with contours extracted by an automatic algorithm that uses block-matching to enforce a crude frame-to-frame similarity condition. This approach allows training data to be obtained in a rapid and relatively painless way, and provides a means of testing whether the deep neural network architecture, which processes only one image at a time, is able nonetheless to embed a priori knowledge corresponding to this additional constraint.

## 2. METHODS

### 2.1. Deep Neural Networks and autoencoders

*2.1.1. Restricted Boltzmann Machines*

The model of Deep Neural Networks proposed in [2] is based on the stacking of Restricted Boltzmann Machines (RBMs). A Restricted Boltzmann Machine is a neural network composed of a layer with visible units and a layer with hidden units, connected through directional links (weights), which are symmetric. The probability of activation of a hidden unit depends on the weighted activations of the units in the visible layer (and vice-versa, since the connections are symmetric).

*2.1.2. Deep architectures*

Training a deep neural network uses a supervised learning strategy based on the stacking of RBMs trained layer per layer from bottom to top. Using deep networks has several advantages. First of all, deep learning (DL) algorithms provide data-driven feature extraction in which the output of each layer gives a representation of input data. Moreover, DL is able to deal with large sets of data. Deep neural networks often give very good results, which explains why they are currently popular in many signal processing applications [3].

## 2.2. Training strategy

### 2.2.1. Learning the relationship between US and contour

Our method is divided into two phases. In the first phase, the network, acting as an autoencoder, is trained to reproduce its input vector. This vector is the concatenation of an ultrasound image and a binary image that represents the contour of the tongue, both reduced to 33 x 30 pixels, resulting in 1980 components, plus one constant input (bias). In the second phase, the network is asked to learn to reconstruct the tongue contour from the ultrasound image only. If we use a network trained on both contour and ultrasound image inputs, it is not obvious that the network will be able to produce a contour image if it lacks one of the inputs. The method described in [1] proposes to estimate the contours using ultrasound images only, under the hypothesis that the representation learned by a network trained on the two kinds of images embeds the relationship between these two kinds of data. The architecture used is called an autoencoder (see [4] [5] [6] and [7] for details). This type of network is trained to find an internal representation (code) of the input data so that it can be precisely reconstructed from this internal representation only.

In our case, if we are able to build an encoder that can generate a hidden coding like the one produced by the combination of ultrasound and contour images, but using ultrasound data only, then the decoder should be able to decode hidden information to reconstruct both ultrasound and contour data. This encoder is obtained in a "translational" manner [1] from the original encoder: the first RBM is replaced by what is called a translational RBM (tRBM, see Figure 1).

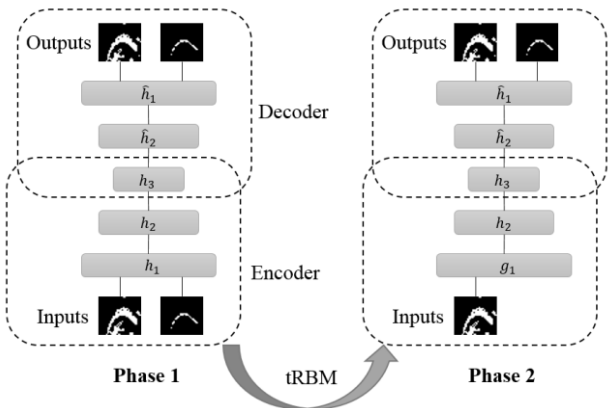

**Figure 1**: The two phases of learning. In the first step, the network learns the relationship between US images and contour. In the second phase, it is able to use this relationship to reconstruct the contour.

In the second phase, we learn only the parameters of the first layer, the others remaining unchanged. In other words, tRBM is trained to produce the same hidden features as the features extracted from the original RBM but without contour inputs. Then, if we use the original autoencoder but replace the first RBM by the tRBM, we can reconstruct a contour image that matches the tongue shape for each reduced ultrasound image of the test database.

### 2.2.2. Initial labeling

For both training and test, we used data from recording sessions described in [8]. We use in our training database an initial contour automatically extracted with an image processing algorithm developed in the Max/MSP software environment that outputs a tongue surface for each ultrasound image. Each ultrasound image is first pre-processed in order to select a region of interest of the relevant portion. On these images, the contour detection is done columnwise, from left to right. For each column, from top to bottom, every white pixel followed by a black pixel is considered a candidate contour point. This implies that several pixels can be selected as candidates. Since only one pixel per column is retained, a decision is made as to which candidate indeed belongs to the contour. This is done by comparing the current image to the previous one, the idea being that if a pixel was part of the previous contour, it or one of its neighbours must belong to the next one. If however no point from the previous contour matches one of the candidates, the selection is based on the neighbouring columns. Using this procedure, we pick up a set of $(x, y)$ coordinates corresponding to the tongue surface contour in each image. These coordinates are then used as the ground truth (referred to as Ref) for the training set of the autoencoder.

## 3. CHOICE OF NETWORK ARCHITECTURE

Each example from the training dataset is presented to the network as an array containing the normalized intensities of the two binary images (1980 pixels + 1 bias). Several hyperparameter sets of the structure were explored (defined in sections 3.1-3.4): the number of layers, the number of unit per layer, the number of epochs for training and the size of "mini-batches", which are subsets of training data, usually of 10 to 100 examples. Our choice of parameters was based on the validation error (root mean squared difference between input and reconstruction) on a 17,000 example dataset (15,000 examples for training and 2,000 for validation).

## 3.1. Deep architectures

Stacking RBMs increases the level of abstraction of the model. However, we must determine the appropriate depth. For this purpose, we tried several architectures with various depths. In our experiments, we fixed 1,000 units per layer, 50 epochs and mini-batches of size 1,000 and tested the performances for a structure with 2, 3 and 4 hidden layers. The lowest validation error was achieved while using 3 hidden layers (see table 1).

**Table 1**: Influence of the number of layers on the validation error.

| Number of hidden layers | Validation error |
|---|---|
| 2 | 0.39 |
| 3 | **0.38** |
| 4 | 0.44 |

## 3.2. Network complexity

In classical machine learning models, we should use more training cases than parameters to avoid overfitting [9]. However, it is common to have a large number of hidden units in deep architectures. For our application, we based our choice of hidden units on the performances of several configurations allowing reasonable computing time, shown in table 2.

**Table 2**: Influence of the number of hidden units on the validation error for the 3 layer model.

| Number of hidden units per layer | Validation error |
|---|---|
| 500 | 0.41 |
| 1000 | 0.38 |
| 2000 | **0.37** |

## 3.3. Use of mini-batches

The use of mini-batches speeds up the algorithm because a weighted update occurs for each mini-batch instead of each example. However, finding an ideal mini-batch size is not straightforward. According to [9], the training set should be divided into mini-batches of 10 to 100 examples. We decided to test tongue contour reconstruction using several mini-batch sizes: 10, 50 and 100 examples per mini-batch.

**Table 3**: influence of mini-batch size on the validation error.

| Mini-batch size | Validation error |
|---|---|
| 10 | 0.65 |
| 50 | 0.53 |
| 100 | **0.38** |
| 200 | 0.40 |

Results showed that for a 3 layer network with 1,000 units per layer, 50 epochs and mini-batches of size 10, the error reached 0.65, while it decreased to 0.38 with mini-batches of size 100 and increases above.

## 3.4 Number of epochs

We used a similar procedure for testing the number of epochs necessary for weight updates. Keeping a reasonable number of epochs is crucial for computation time, and the time vs. performance balance should be considered. We used a 3 hidden layer network with 1,000 hidden units per layer, using mini-batches of size 100, and tested 5, 50 and 250 epoch runs. Using too many epochs degrades the performances. Furthermore, the number of epochs is one of the main bottlenecks for computation time.

**Table 4**: Influence of the number of training epochs on the training error.

| Number of epochs | Validation error |
|---|---|
| 5 | 0.41 |
| 50 | **0.38** |
| 250 | 0.40 |

## 4. RESULTS

### 4.1. Evaluation criteria

During the training stage, we used an autoencoder made of a 3-layer encoder associated with a symmetric decoder, with 2,000 hidden units, mini-batches of size 100 and 50 epochs. The evaluation of the quality of tongue shape reconstruction requires definite criteria and comparison to a reference. Generally speaking, a proper tongue shape is a curve that follows in a realistic manner the lower edge of the bright line appearing on an ultrasound image. It is important to extract the entire visible surface appearing in the ultrasound image, without adding artifacts [10]. In order to evaluate the quality of tongue shapes obtained with the DL method, we trained the network on a 17,000 example database and randomly selected another 50 ultrasound images from the same recording session and same speaker to test the tongue contour extraction. We first compared the contour coordinates obtained with DL to those obtained with manual labelling. However, the set of tongue contour coordinates does not always have the same number of points (see figure 2), so that comparison between coordinates is not straightforward. In [11], a measure is proposed to compare each pixel of a given curve to the nearest pixel (in terms of $L_1$ distance) on the curve it is compared to. This measure, named Mean Sum of

Distances (MSD) (see eq. (1)), provides an evaluation in pixels of the mean distance from a contour $U$ to a contour $V$, even if these curves do not share the same coordinates on the $x$ axis or do not have the same number of points. Contours are defined as a set of $(x, y)$ coordinates: $U$ is a set of 2D points $(u_1, ..., u_n)$ and $V$ is a set of 2D points $(v_1, ..., v_m)$. MSD is defined as followed:

$$MSD(U,V) = \frac{1}{m+n}\left(\sum_{i=1}^{m}\min_{j}|v_i - u_j| + \sum_{i=1}^{n}\min_{j}|u_i - v_j|\right). \quad (1)$$

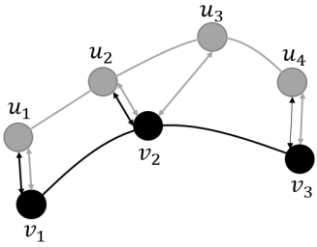

**Figure 2**: Comparing two tongue contours using MSD allows a comparison between two shapes even if some points are missing.

### 4.2. Experiments

Some example contours found using DL are shown in Figure 3. It now remains to compare the various methods used and evaluate the results. In addition to the comparison of the coordinates from DL to manual labelling (Hand), of course, we also want to compare automatically labelled ground truth (Ref) computed in sec. 2.2.2, to manual labelling in order to complete our analysis. Results appear in Table 5.

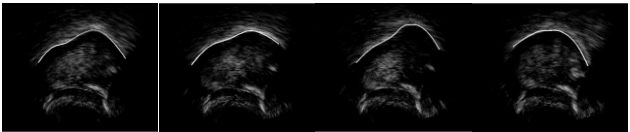

**Figure 3:** Examples of extracted contours for different tongue shapes.

The results show that the contours obtained using DL, Ref, and Hand labelling are of rather similar quality. This implies that the DL autoencoder, despite being shown only one image at a time, was able to achieve results comparable to an algorithm, Ref, that has access to the preceding image in the sequence. This suggests that the DL architecture has embedded a priori structural information stemming from the similarity constraint imposed in the Ref algorithm. This is an interesting result that may open the way to incorporating additional structural cues, for example from a physical 3D tongue model.

We also wished to compare these MSD values to those reported in the literature. In [11], the labelling from EdgeTrak, which uses Snake method (see [12] and [13]), is compared to two manual inputs from two different experts. To compare MSD values in pixels for different resolutions, we converted these values into millimetres using image resolution. Image size was 112.9 x 89.67 mm. The comparison between an expert 1 and an expert 2 gives a MSD of 0.85 mm (2.9 pixels with the conversion 1 px = 0.295 mm), the comparison between expert 1 and EdgeTrak gives a MSD of 0.67 mm, while the comparison between expert 2 and EdgeTrak gives an MSD of 0.86 mm. In [1], after 5 cross-validations, the average MSD computed on 8640 images is 0.73 mm. Our MSD values, computed with the equivalence 1 px = 0.35 mm, are quite similar to these, which allows us to conclude that the results obtained using DL trained with an automatic algorithm are of good quality.

**Table 5**: Average values of MSD for the comparison between Hand and Ref; Hand and DL; and Ref and DL.

|  | Average MSD (mm) |
|---|---|
| Hand vs. Ref | 0.9 |
| Hand vs. DL | 1.0 |
| Ref vs. DL | 0.8 |

### 5. DISCUSSION

The use of a deep autoencoder to automatically extract the contour of the tongue from an ultrasound picture appears to give promising results. The results also show the interest of using automatically extracted contours as ground truth instead of manually labelling large amount of data, which is time consuming. Moreover, since our technique provides performances similar to those of an algorithm that uses temporal information, we can consider that our network was able to learn a new constraint based on its inputs, even if it does not use temporal prior knowledge. The choice of our network structure was adjusted and validated by several performance tests. In the future, providing the algorithm with a variety of learning databases, composed of sentences, words or phonemes pronounced by several speakers and in various modalities (e.g., speech or singing) would be a way to testing the sensitivity of the algorithm to variations in experimental conditions.

### 7. ACKNOWLEDGEMENTS

This work is funded by the European Commission via the i-Treasures project (FP7-ICT-2011-9-600676-i-Treasures).
We are also grateful to Cécile Abdo for the algorithm she developed on the Max/MSP software environment.

# 6. REFERENCES

[1] Fasel, I., Berry, J. 2010. Deep Belief Networks for Real-Time Extraction of Tongue Contours from Ultrasound During Speech. *20th International Conference on Pattern Recognition* Istanbul, IEEE, 1493-1496.

[2] Hinton G. E., Osindero, S. 2006, A fast learning algorithm for deep belief nets. *Neural Computation,* 18, 1527-1554.

[3] Yu, D., Deng, L. 2011. Deep Learning and Its Applications to Signal and Information Processing. *IEEE Signal Processing Magazine,* 28, 245-254.

[4] Larochelle, H., Bengio, Y., Louradour, J., Lamblin, P. 2009. Exploring Strategies for Training Deep Neural Networks. *Journal of Machine Learning Research,* 10, 1-40.

[5] Bengio, Y. 2009. Learning Deep Architectures for AI. *Foundations and Trends in Machine Learning,* 2, 1-127.

[6] Vincent, P. Larochelle, H. Lajoie, I., Bengio, Y., Manzagol, P-A. 2010. Stacked Denoising Autoencoders: Learning Useful Representations in a Deep Network with a Local Denoising Criterion. *Journal of Machine Learning Research,* 11, 3371-3408.

[7] Arnold, L., Rebecchi, S., Chevallier, S., Paugam-Moisy, L. 2011. An introduction to deep-learning. *Advances in Computational Intelligence and Machine Learning, ESANN'2011,* Bruges, 477-488.

[8] Cai, J., Hueber, T., Manitsaris, S., Roussel, P., Crevier-Buchman, L., Stone, M., Pillot-Loiseau, C., Chollet, G., Dreyfus, G., Denby, B. 2013. Vocal Tract Imaging System for Post-Laryngectomy Voice Replacement. *International IEEE Instrumentation and Measurement Technology Conference*, Minneapolis, MN.

[9] Hinton, G. E. 2012. A Practical Guide to Training Restricted Boltzmann Machines. In: Montavon, G., Orr, G. B., Müller, K-R. (eds.), *Neural Networks: Tricks of the Trade (2nd ed.)*, Springer, 599-619.

[10] Stone, M. 2005. A Guide to Analysing Tongue Motion from Ultrasound Images. *Clinical Linguistics and Phonetics,* 19, 455-502.

[11] Li, M., Kambhamettu, R., Stone, M. 2005. Automatic Contour Tracking in Ultrasound Images. *Clin. Linguist. Phon.*, 19, 545-554.

[12] Kass, M., Witkin, A., Terzopoulos, D. 1988. Snakes: Active contour models. *International Journal of Computer Vision,* 1, 321-331.

[13] Akgul, Y. S., Kambhamettu, C., Stone, M. 1999. Automatic Extraction and Tracking of The Tongue Contours. *IEEE Transactions on Medical Imaging*, 18, 1035-1045.